# Vision-enabled detection of safety helmet compliance in construction zones

**Tri Nhut Do[1, 2*], Ba Loc Pham[3]**

[1]*University of Information Technology, Quarter 34, Linh Xuan Ward, Ho Chi Minh City, Vietnam*
[2]*Vietnam National University - Ho Chi Minh City, Quarter 34, Linh Xuan Ward, Ho Chi Minh City, Vietnam*
[3]*Thu Dau Mot University, 6 Tran Van On Street, Phu Loi Ward, Ho Chi Minh City, Vietnam*

Received 24 December 2024; revised 15 January 2025; accepted 10 June 2025

***Abstract:***

**In the rapidly evolving field of construction management, worker safety remains a top priority. This paper introduces an innovative vision-based system for real-time detection of helmet compliance, specifically designed for construction sites, utilising advanced computer vision techniques and machine learning algorithms within the YOLO (you only look once) framework. Our system leverages high-resolution video feeds from strategically positioned cameras to monitor adherence to safety regulations concerning helmet usage. By employing deep learning methodologies, the system effectively identifies individuals not wearing helmets, thereby significantly mitigating the risk of head injuries among workers. Our training and validation results reveal an impressive precision exceeding 97% at mAP@0.5 for both helmeted and non-helmeted individuals. Furthermore, our experiments demonstrate exceptional detection accuracy, showcasing the system's resilience under varying lighting conditions and diverse worker movements. The consistent decrease in loss and improvement in metrics throughout training validate the effectiveness of the YOLOv8 model in enhancing recognition performance. The implications of this research extend beyond mere regulatory compliance, opening avenues for innovative applications in occupational safety management. This study highlights the critical role of technology in protecting lives and lays the groundwork for future advancements in smart construction environments.**

## 1. Introduction

The construction industry is inherently hazardous, with a high incidence of accidents and injuries. According to the U.S. Bureau of Labor Statistics (2023) [1], helmet-related head injuries are among the leading causes of construction site fatalities. As awareness of safety increases, regulatory bodies enforce stringent compliance measures to mitigate risks, emphasising the need for effective monitoring solutions. According to the International Labour Organization (ILO), construction is one of the most dangerous industries, with an average fatality rate of 12.4 per 100,000 workers per year [2]. Therefore, the use of personal protective equipment (PPE) such as helmets (or hard hats) is a simple and effective way to reduce the risk of head injuries and ensure safety at construction sites. Traditional safety measures, such as manual inspections and safety audits, often fall short due to human error, oversight, and limited resources. In light of these challenges, the adoption of technology-driven safety solutions has gained traction. This paper introduces a vision-based helmet detection system specifically designed for construction sites, utilising advanced computer vision and deep learning techniques to enhance worker safety and compliance in real-time.

The intersection of technology and occupational safety has been the subject of extensive research. Several studies have explored the application of computer vision in monitoring safety compliance. For instance, Y.J. Zhang, et al. (2022) [3] developed a video analysis framework that utilised image processing techniques to detect PPE compliance on construction sites. Their work highlighted the potential of automated systems in reducing manual oversight, though it lacked real-time processing capabilities, which are critical in dynamic environments. This research not only underscores the importance of protective gear but also advances the application of deep learning in occupational safety. The improved YOLOv5s model represents a significant step forward in developing intelligent detection systems that promote a safer working environment, paving the way for further innovations in safety management technology.

*Corresponding author: Email: trinhutdo@uit.edu.vn

Further advancements were made by Y. Qian, et al. (2023) [4], who employed convolutional neural networks (CNNs) to implement real-time detection of helmet usage. Their methodology achieved high accuracy rates but faced challenges in working under variable lighting conditions and occlusions - a common occurrence in construction environments. Additionally, S. Islam, et al. (2021) [5] presented a comprehensive survey of deep learning applications in construction safety, emphasising the necessity of robust detection systems that can adapt to numerous scenarios found on construction sites. A. Hayat, et al. (2022) [6] contributed significantly to the field of construction safety by demonstrating the potential of deep learning technologies to transform safety monitoring practices. The integration of an automatic helmet detection system heralds a new era in occupational safety, promising reduced accident rates and improved working conditions for construction personnel. The findings advocate for broader adoption of such intelligent systems across the industry, ultimately paving the way for enhanced safety standards and regulations. Furthermore, the study emphasises the practical implications of the system, including the potential for real-time alerts, data analytics for safety audits, and integration with existing site management applications. By automating helmet detection, the system not only enhances compliance with safety regulations but also fosters a proactive safety culture within construction environments. Through this research, we transcend previous studies by merging state-of-the-art machine learning models with practical applications, ultimately contributing to the growing discourse on enhancing safety management in the construction industry.

To enhance the monitoring of helmet-wearing on construction sites to minimise head injuries for workers, we present in this paper the design and practical application of a vision-based system to automatically detect construction workers not wearing helmets. These technology-based solutions [7-9] need to be implemented in conjunction with enforcement and education strategies, as well as providing incentives for workers to wear helmets, to most effectively achieve the goal of increasing helmet-wearing rates on construction sites. Therefore, this paper seeks to address these limitations by proposing a comprehensive vision-based helmet detection system that not only performs effectively across diverse conditions but also supports real-time alerts. By analysing and learning from a wide array of datasets, our system aims to create a safer construction workplace, reducing the burden on safety personnel and enhancing compliance efforts. To ensure that the system accurately detects people without helmets on construction sites, we employ two recognition algorithms: face recognition and helmet recognition. These algorithms are all based on the YOLO architecture. YOLO is a real-time object detection algorithm that detects objects in a single step, unlike other methods that require multiple steps. YOLO is designed to detect objects in a single image (a frame) and a sequence of multiple images (video frames), including faces, pedestrians, cars, bicycles, etc. The recognition results of both algorithms are frame coordinates (in pixels) on the same displayed image. A pair of coordinates is defined as extracted from the face recognition algorithm and the helmet recognition algorithm. Finally, a compatibility check algorithm in a pair of coordinates is proposed and applied to eliminate. The single coordinates extracted from the face recognition algorithm that cannot create compatibility are defined as people (objects) without helmets. This is also the final result of the system for detecting workers not wearing safety helmets on construction sites proposed in this article. This result will be displayed at the end.

This article is structured as follows. Section 2 describes the background review, while Section 3 details the design of the proposed system for real-time application, including the design of the hardware and software components. Section 4 presents some experimental results that demonstrate the accuracy of the helmet detection algorithm using the proposed system. Finally, Section 5 concludes the paper and suggests some future directions for improvement.

## 2. Background review of recognition algorithms

In this section, we summarise the fundamental issues of the two algorithms applied in this proposed system. First, the YOLO algorithm extracts features from the input image using a CNN [10]. These features are used to detect objects in the image. Next, YOLO uses a single neural network to predict bounding boxes and class probabilities for each object in the image. Finally, to remove overlapping bounding boxes, YOLO applies non-maximum suppression, which reduces the number of detected objects. The YOLO face detection algorithm is an extension of the YOLO architecture specifically designed to detect human faces in images and videos. The algorithm consists of three main stages:

Stage 1: Feature extraction. The first phase involves extracting features from the input image using a CNN. The CNN is trained on a dataset of images containing human faces and images without human faces.

Stage 2: Object detection. The second phase involves detecting faces in the input image using the extracted features. The algorithm uses a technique called non-maximum suppression to eliminate overlapping bounding boxes and reduce the number of detected faces.

Stage 3: Post-processing. The final phase involves post-processing the detected faces to improve the algorithm's accuracy. This phase involves applying various techniques such as face landmark detection, face alignment, and face recognition.

Therefore, the proposed YOLO-based helmet detection algorithm in this paper is also an extension of the YOLO architecture specifically designed to detect helmets in images and videos. The algorithm also consists of three main stages:

Stage 1: Feature extraction. In the first stage, the CNN is trained on a dataset of helmet images and non-helmet images.

Stage 2: Object detection. The second stage involves detecting helmet objects in the input image using the extracted features.

Stage 3: Post-processing. This stage involves applying various techniques such as object classification, pose estimation, and tracking.

## 3. System design

Detecting individuals with and without helmets on a construction site is crucial for ensuring safety and compliance with regulations. Various systems and technologies that can be implemented for this purpose [11-14] are illustrated in Fig. 1, ranging from simple manual checks to advanced automated systems.

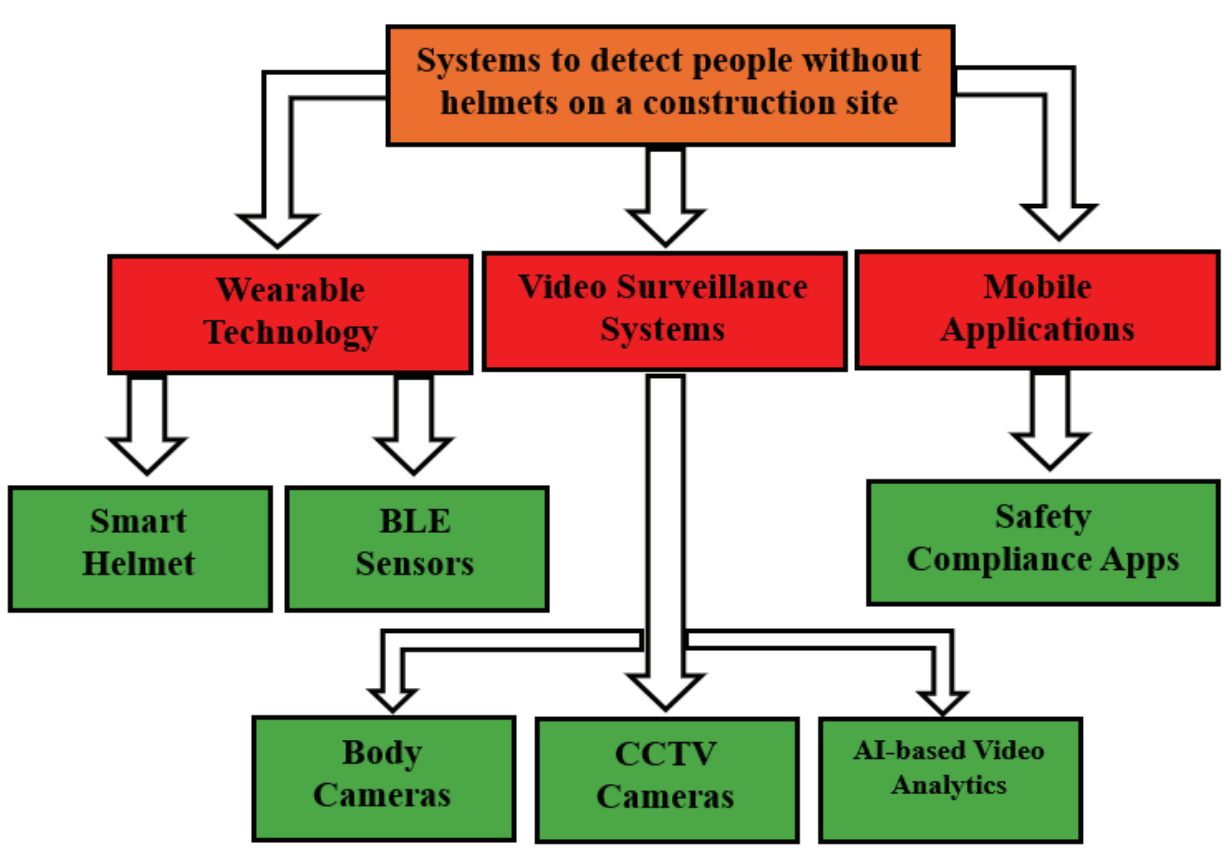

**Fig. 1. Detection system types.**

Here are some options:

*Mobile applications:*

Safety compliance apps: Develop or implement mobile apps that allow supervisors to capture images of workers on-site. The app can utilise image recognition technology to verify compliance with PPE [15-23] regulations. The approaches mentioned indicate that using a single evaluation measure is insufficient, yet they do not elaborate on which specific metrics should be included or how they should be weighted. Therefore, our evaluation framework integrates standard metrics - accuracy, precision, recall, F1-score, and mean Average Precision (mAP) - with confidence threshold optimisation to provide a practical, deployment-focused assessment of YOLOv8's performance, addressing the unique challenges of construction site monitoring. Moreover, approaches [15-24] implemented a real-time multi-threaded inference engine with a frame rate of 30 FPS, but they do not indicate the latency or computational requirements of the system in practice. In this study, an additional analysis of the system's requirements and potential real-world limitations would be beneficial.

*Wearable technology:*

Smart Helmets: Utilise smart helmets equipped with sensors that detect if a helmet is being worn correctly. These helmets can also include communication features to alert supervisors if someone is not wearing their helmet.

Bluetooth Low Energy (BLE) Sensors: Employ BLE beacons on helmets that can communicate with smartphones or receivers. Only individuals wearing helmets will be recognised as being present on-site.

*Video surveillance systems:*

Body cameras: Workers can wear body cameras that have built-in image recognition capabilities to alert them when they are near someone without a helmet.

CCTV cameras: Deploy CCTV cameras at strategic locations around the site. These can be monitored in real-time or recorded for later review.

AI-based video analytics: Utilise AI algorithms such as ML, DL, and YOLO that can analyse video feeds to detect individuals without helmets. These systems can provide real-time alerts to supervisors.

The block diagram shown in Fig. 2 represents a system that effectively monitors individuals who do not wear helmets using video input from a camera. The system is structured sequentially, comprising several key components that work together to identify people not wearing helmets, trigger alarms, and alert supervisors. The system begins with a video or camera input block, where the camera captures real-time footage of the monitored area. This input can come from various types of cameras, including stationary, PTZ (pan-tilt-zoom), or wearable cameras. The purpose of this block is to continuously stream video data that serves as the primary source for real-time analysis.

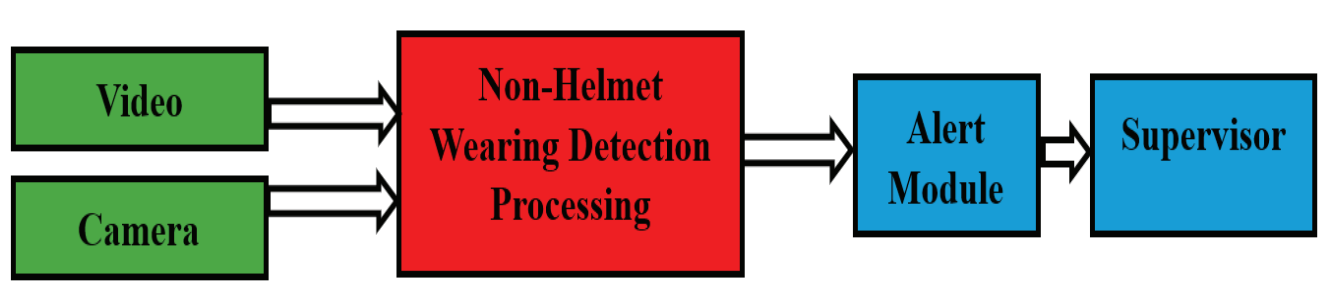

**Fig. 2. Non-helmet wearing detection system.**

The second block is the non-helmet wearing detection processing. This block acts as the core processing unit of the system. It utilises computer vision and machine learning algorithms in the YOLO architecture to analyse the incoming video feed. The next block is the alert module. Once a non-helmet wearing is detected, the information is channelled into the alert module, which acts as the decision-making driver. This module can operate on different criteria, such as the severity of the incident or the time elapsed since detection. Based on the detection outcomes, the alert module determines the appropriate response. It may also evaluate whether to send a notification to site supervisors, safety officers or emergency response teams, depending on the configuration of the system.

The final block is the supervisor block as output. This block serves as the management output component of the system. It is triggered by the alert module when non-helmet wearing is confirmed.

In summary, the block diagram of the non-helmet wearing detection system illustrates a streamlined process where video captured from a camera is analysed in real-time to detect individuals who do not wear helmets, prompting an alert. This integrated approach enhances safety and response times on construction sites.

### *3.1. Helmet and non-helmet wearing detection processing design*

In this section, the authors propose the design of a system to identify individuals with and without helmets on construction sites, based on computer vision techniques and the YOLO architecture algorithm version 8. This proposed design system features an overview architecture with two output classes: helmet and no_helmet, as shown in Fig. 3. In this system, YOLO version 8 architecture is applied as a deep learning framework for detection processing to identify individuals without helmets, with a block diagram of steps as shown in Fig. 4.

In Fig. 4, the YOLO version 8 architecture is described in detail as follows:

Input: This is the input data of the model, a 3-channel RGB colour image with a size of 640x640 pixels.

Backbone (CSPNet+SPP): This is to extract features from input images.

CSPNet (Cross stage partial network): This component reduces the number of computations while maintaining the efficiency of feature extraction. In addition, this is to divide and combine the computational branches in order to reduce the computational cost.

SPP (Spatial pyramid pooling): This component summarises global information and increase the ability to recognise objects of different sizes. Moreover, this is to use pooling layers with different sizes (e.g. 1×1, 5×5, 9×9).

Neck (Path aggregation network - PANet): This component increases the ability to aggregate information from different feature levels (deep and shallow features) and to perform connections from different feature maps, from backbone to head. Finally, this is to improve object detection performance at small and large sizes.

Head: This component is responsible for generating the output. For YOLOv8, it utilises multi-scale detection levels (80×80, 40×40, 20×20). Each level corresponds to different object sizes: 80×80: For small objects; 40×40: For medium objects; 20×20: For large objects.

Output (BBox+Class): This is to give the final result, including the detected object location in the image and the detected object type.

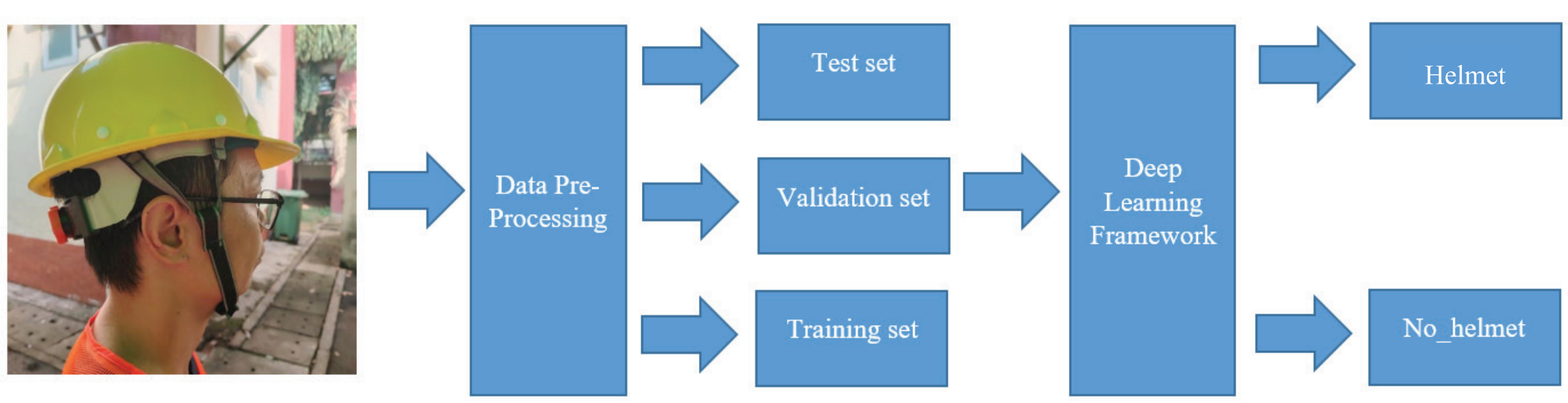

**Fig. 3. Overview architecture for the proposed system based on deep learning.**

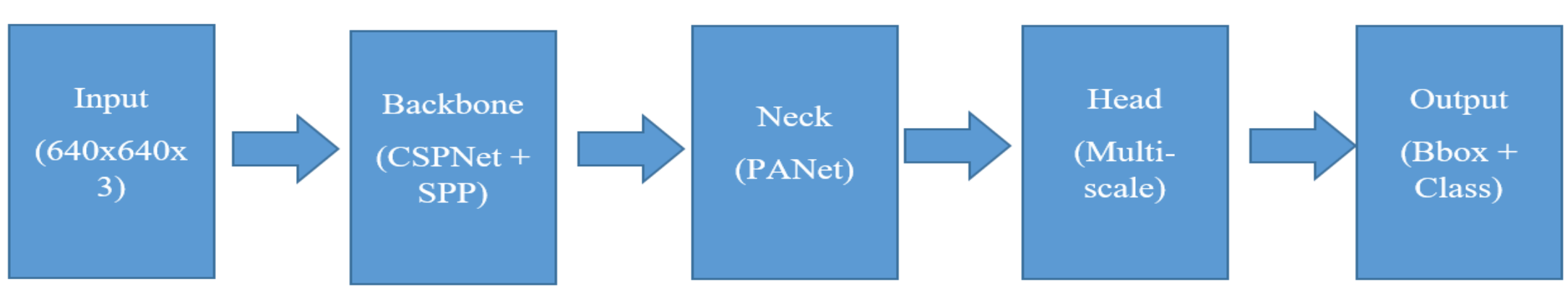

**Fig. 4. YOLO version 8 architecture.**

BBox (Bounding Box): This is the bounding box coordinates (x, y, width, height) surrounding the object.

Class: This is the detected object label (e.g.: helmet, no helmet).

*3.1.1. Data capturing and labelling*

The images used for training were captured at angles that can be recognised by the camera when workers are on the construction site. These training images were neither too blurry nor too distant from the human eye. We use a publicly available dataset from Kaggle, such as the YOLO Helmet/Head Helmet Recognition dataset, consisting of 7035 images, including 2107 images with helmets, 3912 images without helmets, and 1016 images with and without helmets. Additionally, there are 1254 selfies taken with an Honor X9b phone, including 1234 photos with a helmet and 20 pictures without a helmet. The total dataset consists of 8289 photos.

MakeSense.ai tool was used to label all images in the dataset. Firstly, duplicate images are removed after all images are uploaded. The total number of non-duplicate photos is displayed on the screen. Secondly, due to different image resolutions, the image sizes are changed so that at least one side is 640 pixels, adapting to the pixel resolution of the YOLO framework. Thirdly, the dataset is divided into 2 parts including 75% (6210 images) for training and 25% (2079 images) for testing. Finally, the dataset used for training has 2 classes: helmet and no_helmet.

*3.1.2. Training and validating*

The diverse YOLO neural network architectures are employed for training, while all other implementations are executed utilising PyTorch [10, 24].

The processing is performed according to the algorithm flowchart as shown in Fig. 5.

During training, the parameters are tuned simultaneously with increasing epochs until the evaluation metrics are optimally improved. Finally, we obtain the best-trained system.

To evaluate the performance of the YOLO version 8 algorithm, we utilise five key metrics: accuracy, precision, recall, F1 score, and mAP. These performance indicators are derived from the four components of the confusion matrix, which include True Positive (TP, the number of instances correctly predicted as positive), False Positive (FP, the number of instances incorrectly predicted as positive), True Negative (TN, the number of instances correctly predicted as negative), and False Negative (FN, the number of instances incorrectly predicted as negative). Specifically, TP indicates the successful identification of individuals wearing safety helmets, while FP denotes the erroneous classification of individuals as helmet wearers. Conversely, TN reflects the accurate identification of people not wearing helmets, and FN highlights instances where individuals without helmets are incorrectly identified. The mathematical formulations for accuracy, precision, recall, F1 score, and mAP are detailed in Eqs. (1) to (5). Notably, the F1 score serves as the harmonic mean of precision and recall, providing a balanced measure of the algorithm's effectiveness.

$$\text{Accuracy} = (TP + TN)/(TP + FP + TN + FN) \quad (1)$$

$$\text{Precision} = TP/(TP + FP) \quad (2)$$

$$\text{Recall} = TP/(TP + FN) \quad (3)$$

$$\text{F1 score} = 2 * ((\text{Precision} * \text{Recall}))/((\text{Precision} + \text{Recall})) \quad (4)$$

$$\text{mAP} = \frac{\sum_{c=1}^{C} \text{Average Precision } (c)}{C} \quad (5)$$

where C is the total number of output classes. In this study, C=2 (helmet and no-helmet).

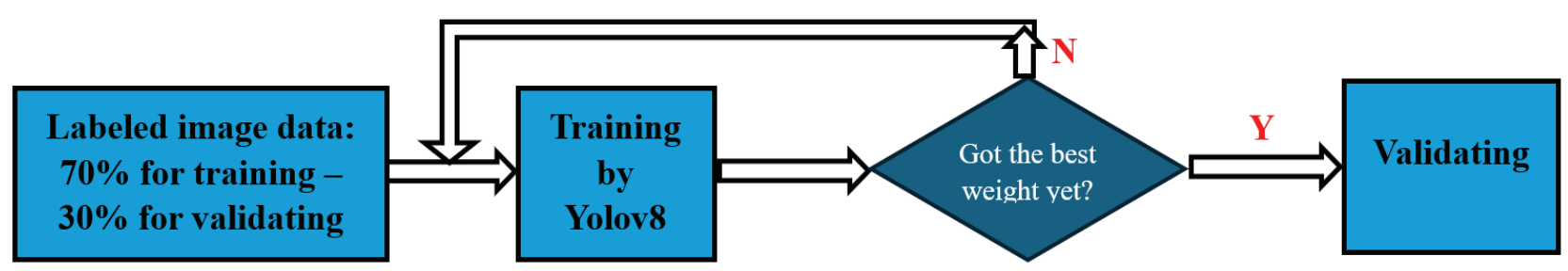

Fig. 5. Detection algorithm flowchart.

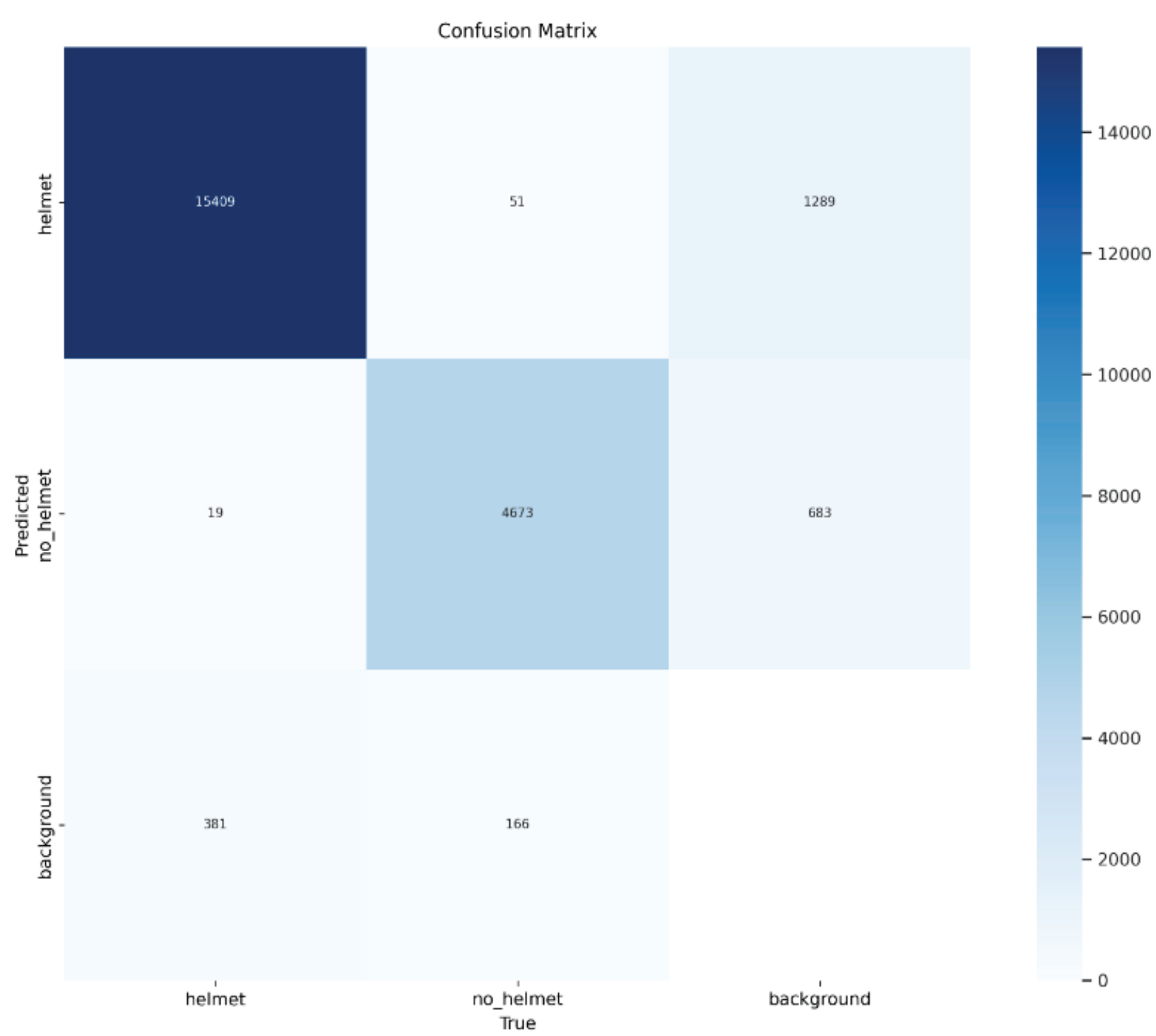

Fig. 6. Confusion matrix.

The quality of the proposed system is evaluated after training by the above metrics. Firstly, the confusion matrix represents the actual versus predicted objects for a comprehensive analysis of the model's accuracy and error types as shown in Fig. 6.

Next, given that the proposed system in this paper is a binary task, the F1-Confidence graph is drawn as shown in Fig. 7 to evaluate the correlation between the F1 score and confidence threshold. The curve illustrates how the F1 score changes as the confidence threshold is adjusted, allowing for the identification of the optimal threshold of 0.444, which maximises the F1 score of 0.94 for the proposed system.

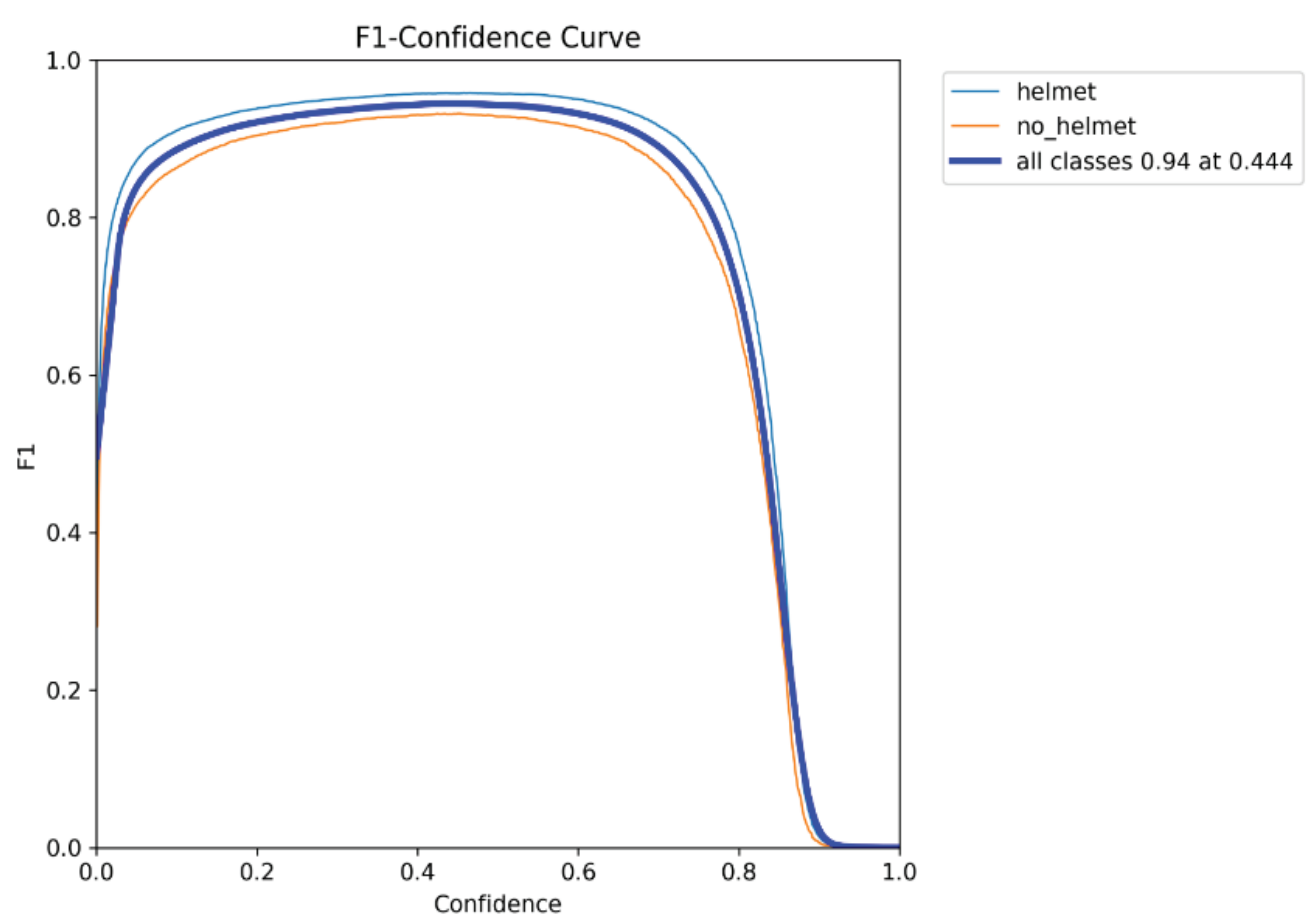

**Fig. 7. F1-Confidence curve.**

Similarly, the Precision-Confidence graph in Fig. 8 evaluates how the precision of the model changes with varying confidence levels, aiding in the selection of optimal thresholds. The curve shows that the threshold at 0.911 for all classes reaches 1.00.

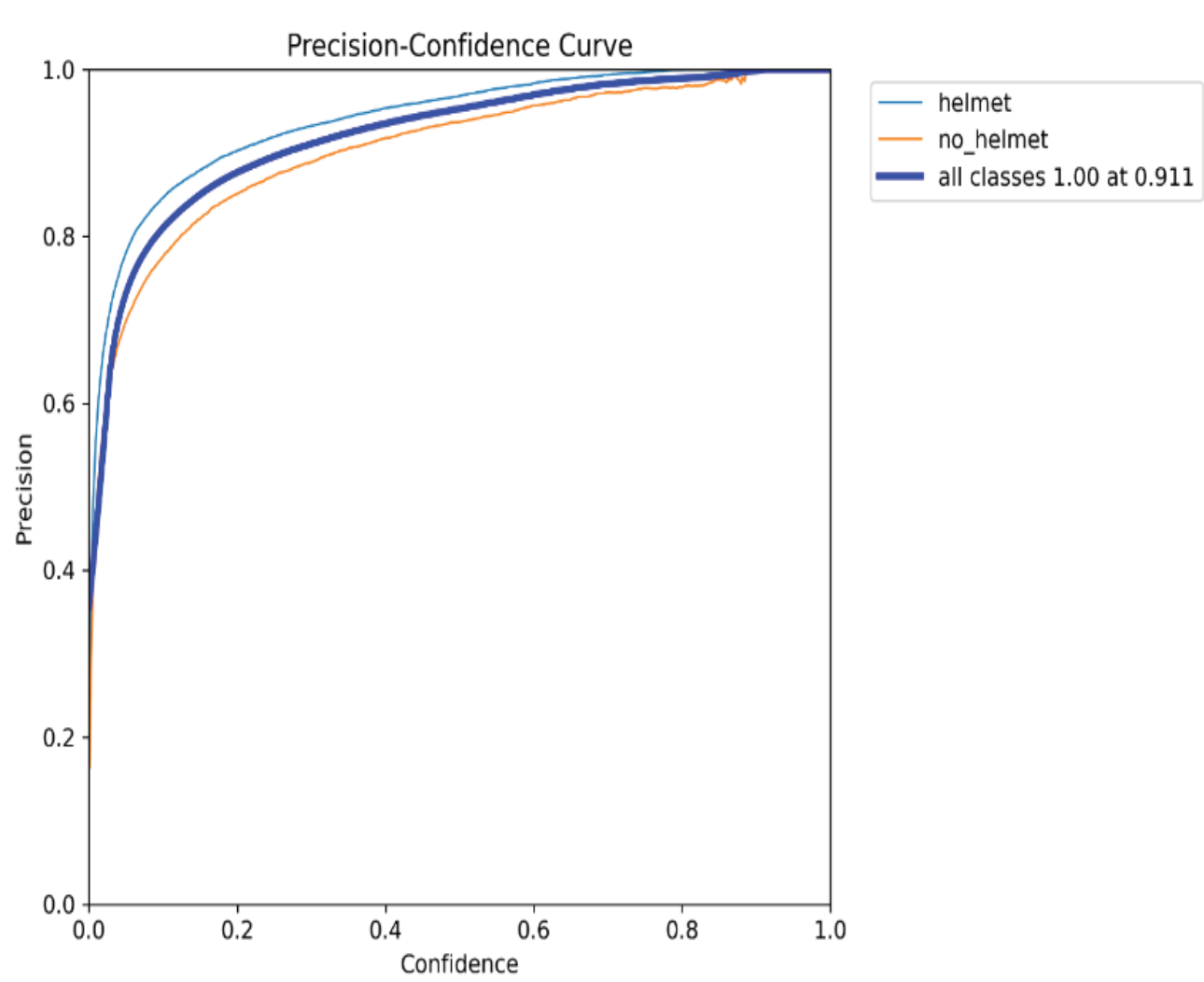

**Fig. 8. Precision-confidence curve.**

Moreover, the Precision-Recall graph is drawn as shown in Fig. 9 to evaluate the trade-off between precision (the ratio of true positive predictions to the total predicted positives) and recall (the ratio of true positive predictions to the total actual positives) at various threshold settings. The curve helps us understand the effectiveness of classification models (the proposed system) in real-world applications.

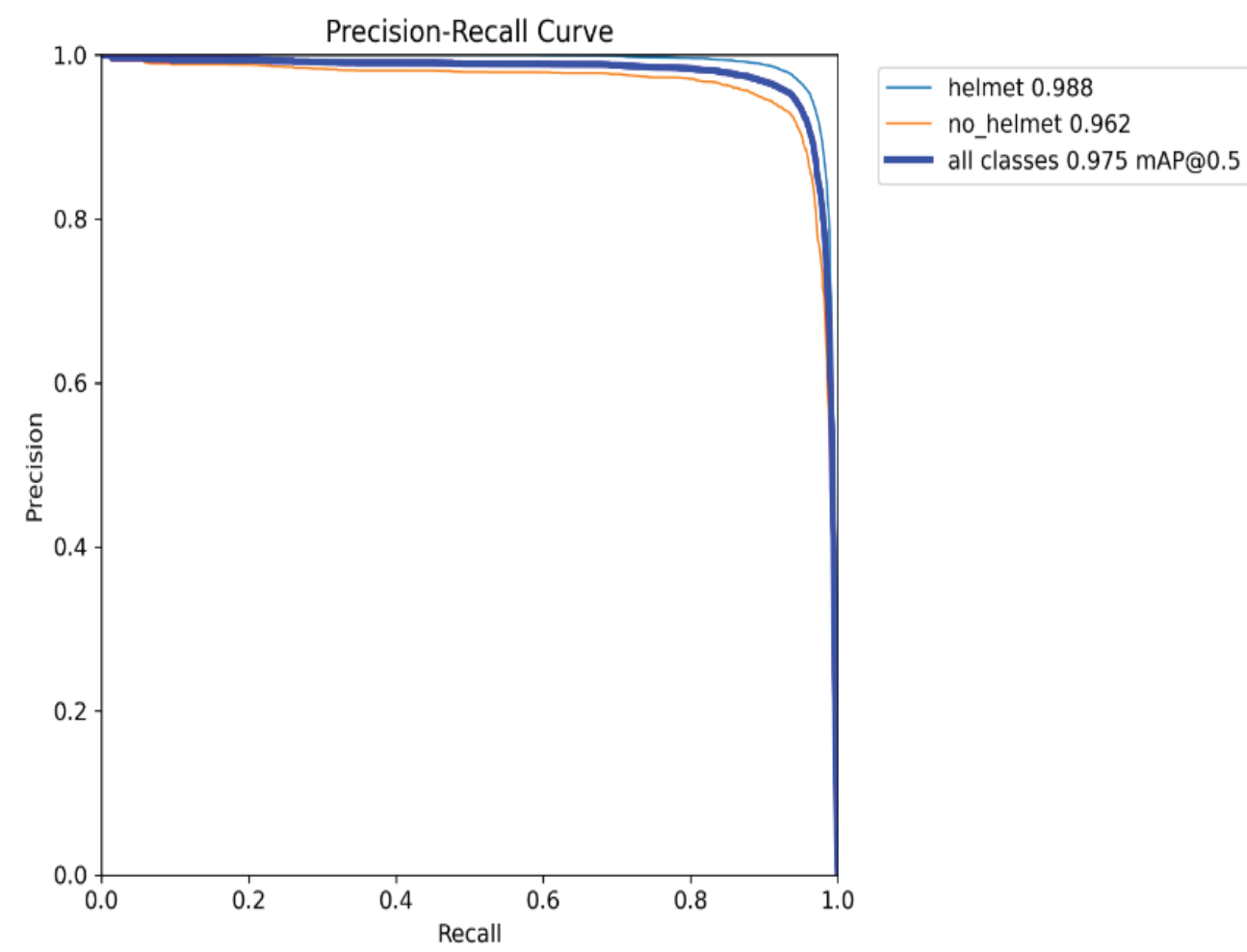

**Fig. 9. Precision-Recall curve.**

Finally, the loss and metrics graphs are shown in Fig. 10. Overall, the results on the graph indicate that the YOLOv8 model is being trained effectively, with a gradual decrease in the loss functions and an increase in the performance metrics. This is a positive sign, suggesting that the model is increasingly suitable for solving the problem of recognising individuals wearing and not wearing hard hats.

The loss values (both train and val) decrease over the epochs, indicating that the model is learning and improving during both the training and testing phases. The loss graphs include the following:

train/box_loss and val/box_loss: This is the loss associated with predicting the location (bounding box) of the objects. The decrease in this loss function during training indicates that the model is becoming increasingly accurate at predicting locations.

train/cls_loss and val/cls_loss: This loss relates to predicting the class of the objects. The decrease in this loss function indicates that the model is improving in classification.

train/dfl_loss and val/dfl_loss: This loss pertains to predicting the distribution focal loss, which is used to enhance the accuracy of the bounding box prediction. The decrease in this loss function is a positive sign.

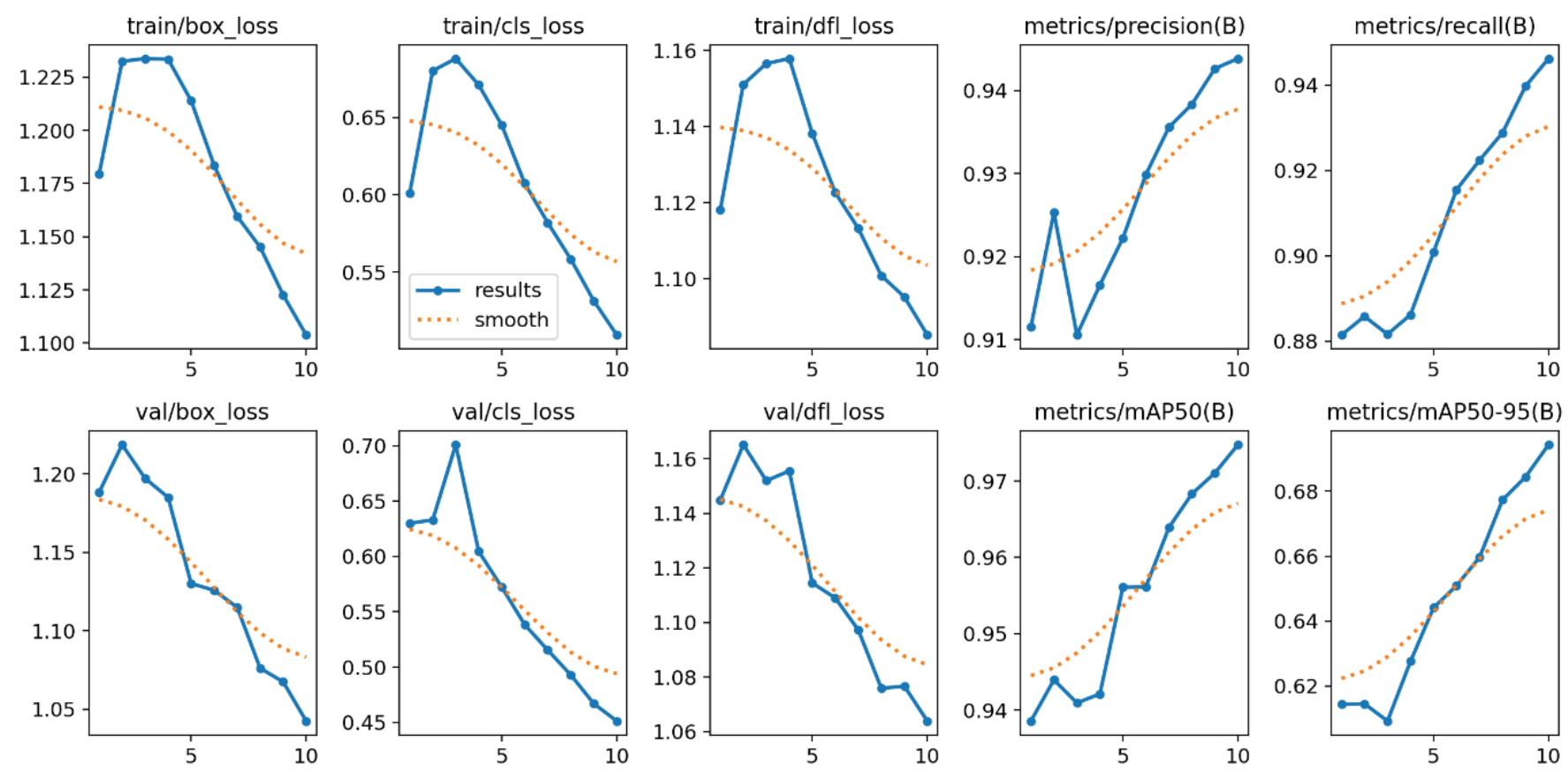

**Fig. 10. The loss and metrics graphs.**

Metrics such as precision, recall, and mAP all increased, demonstrating that the YOLOv8 model learns well and improves overall performance. The metrics graphs include the following:

Metrics/precision(B): Precision measures the proportion of correct predictions among the model's predictions. Increasing precision indicates that the model is reducing incorrect predictions.

Metrics/recall(B): Recall measures the ability to correctly detect objects. The increase in recall indicates that the model is improving at detecting individuals wearing and not wearing hard hats.

Metrics/mAP50(B): mAP at the Intersection over Union (IOU) threshold =50%. A steady increase in mAP50 clearly indicates improved model performance.

Metrics/mAP50-95(B): Mean mAP at IOU thresholds from 50 to 95%. The steady improvement of this metric over epochs indicates that the model is becoming more accurate, even at strict IOU thresholds.

### *3.2. Evaluation for validating*

In this section, we evaluated YOLOv8's performance using five key metrics -accuracy, precision, recall, F1 score, and mAP - derived from the confusion matrix (TP, FP, TN, FN). For our binary helmet detection task, YOLOv8 achieved an F1 score of 0.94 at a confidence threshold of 0.444 (Fig. 7) and a precision of 1.00 at 0.911 (Fig. 8), reflecting its effectiveness in identifying helmeted and non-helmeted individuals under varying conditions. These results, optimised via confidence threshold analysis, underscore YOLOv8's capability to balance precision and recall while ensuring high reliability - key requirements for real-time safety monitoring in construction environments. We selected YOLOv8 initially for its well-documented strengths in speed (critical for on-site deployment) and robustness (e.g., via advanced feature extraction), which align with our system's operational demands.

## 4. Experimental results and evaluation

Creating effective experimental scenarios for a vision-based system to detect individuals without helmets on construction sites requires careful planning and consideration of various factors, including environmental conditions, the diversity of personnel, and the detection system's technological capabilities.

### *4.1. Experimental scenarios and results*

The proposed system was tested across different construction sites (civil construction, building construction, and electrical construction) to understand how these factors affect detection accuracy. All scenarios were conducted outdoors at real construction sites with active workers. This approach helps us monitor and evaluate the system's performance in real-world conditions, where movement patterns and ambient noise vary. The experimental results are illustrated in Fig. 11.

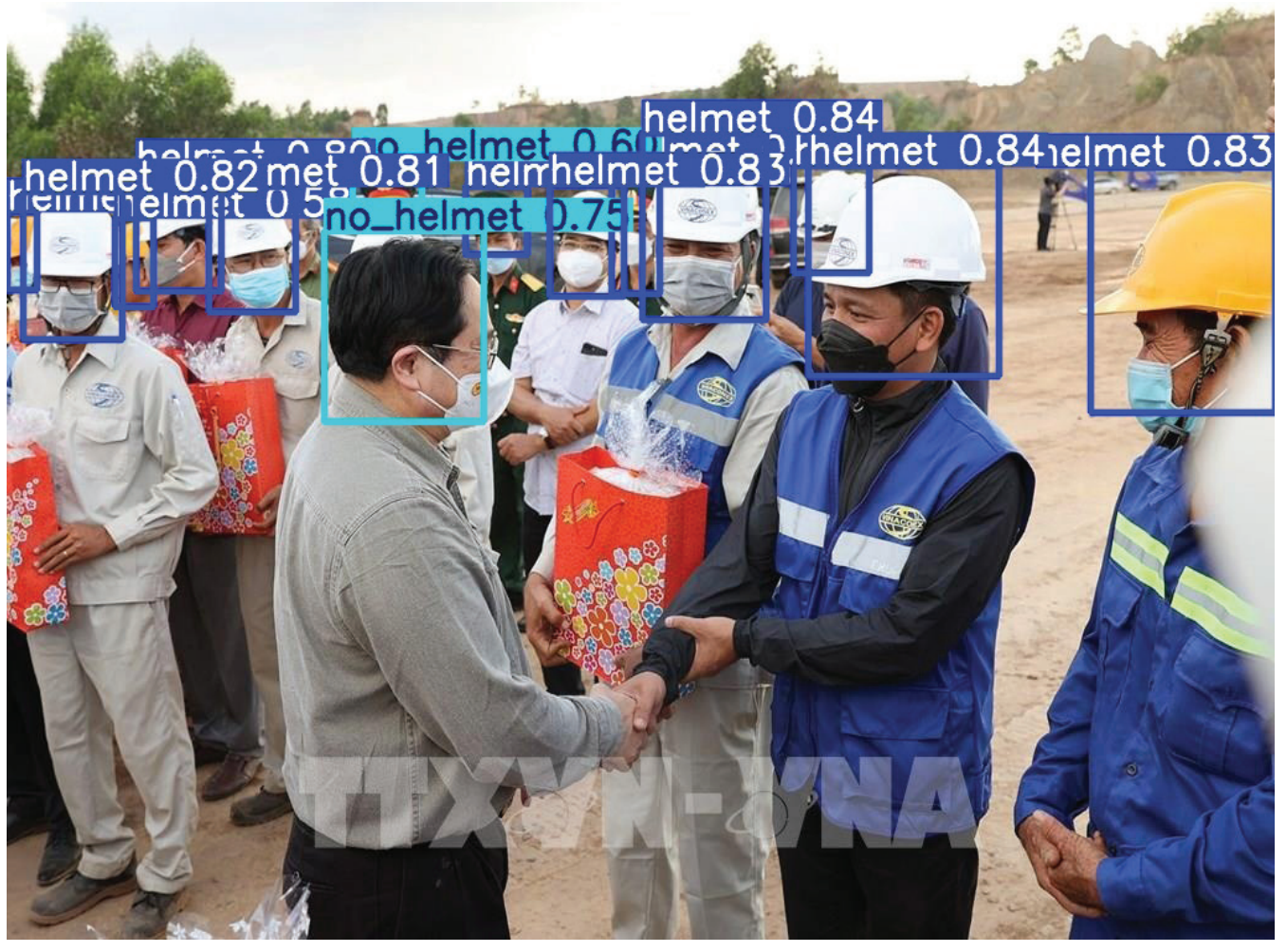

**Fig. 11. Helmets detected by the proposed system on an outdoor construction site.**

Moreover, all experiments were conducted in scenarios where groups of workers moved together, varying the composition of helmeted and non-helmeted individuals. They engaged in various construction tasks such as lifting, assembling, and operating machinery. The experimental results are illustrated in Fig. 12. This approach helps us analyse how well the system tracks dynamic changes in the crowd and determine how these activities affect helmet detection performance.

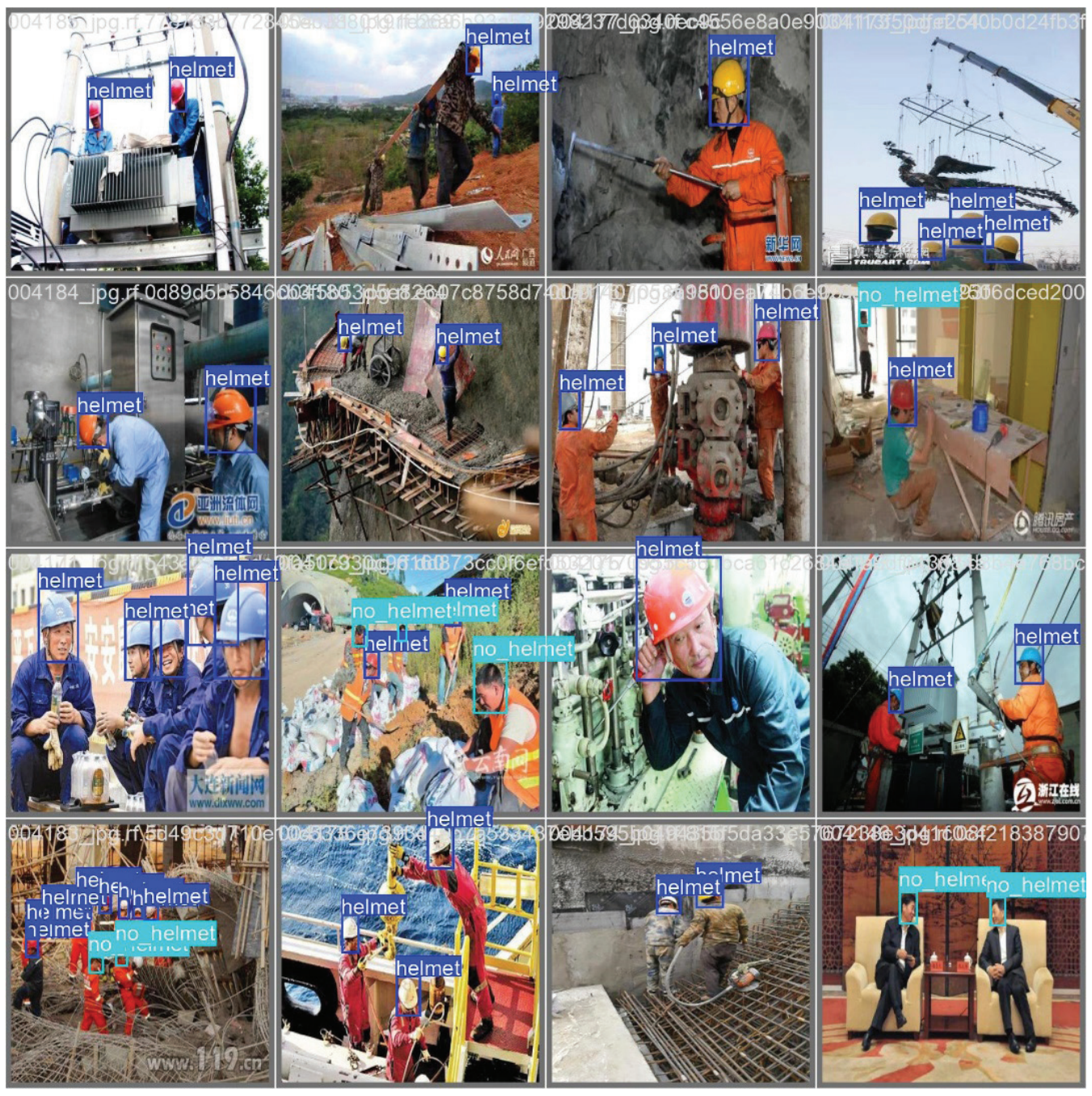

**Fig. 12. Helmets detected by the proposed system on an outdoor construction site.**

Each of these scenarios can provide valuable insights into a vision-based helmet detection system's performance, strengths, and weaknesses, leading to eventual improvements in worker safety management on construction sites.

### *4.2. Experimental evaluation across varying image densities*

In this section, we conducted a benchmark to evaluate our system's performance across varying image densities, addressing the important need for benchmarking our methodology in this context. Our benchmarking strategy specifically involves the Density Spectrum Coverage, which helps us evaluate our methodology using a meticulously curated dataset covering a broad spectrum of image densities. This will include the following:

Low density images: Our proposed system was tested on a dataset of images with a range of densities of less than 10 people. The results are presented in Fig. 12. The system proposed accurately identified individuals without helmets, with almost 100% accuracy.

Medium density images: Our proposed system was tested on a dataset of images with a range of densities between 10 and 20 people. The results are presented in Fig. 11. The proposed system accurately identified individuals without helmets, with almost 100% accuracy.

High density images: Challenging scenarios with testing our approach to crowding and occlusion. Our proposed system was tested on a dataset of images with a range of densities of more than 20 people. Results will be better with subjects closer to the camera's pinhole and closer to the centre of the captured image. The maximum density in our experimental results is 30 people. The results are presented in Fig. 13. The proposed system accurately identified individuals without helmets with nearly 100% accuracy.

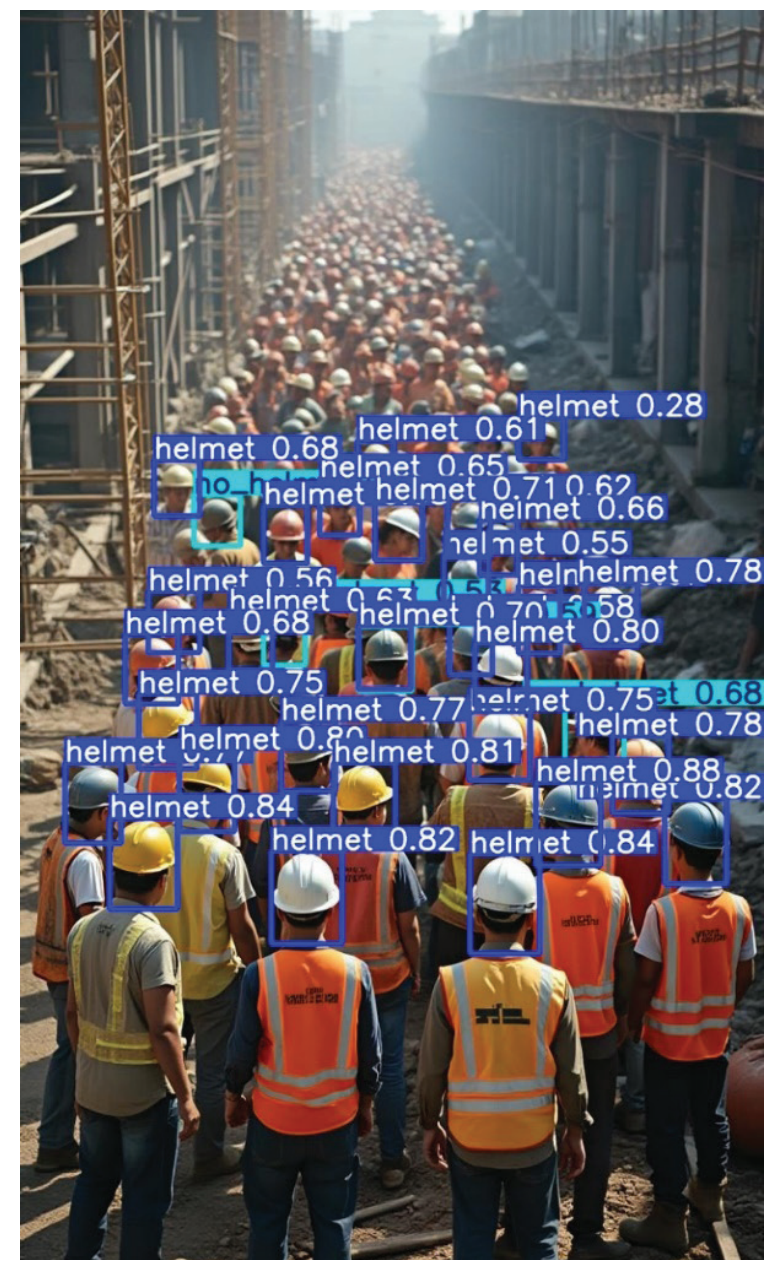

**Fig. 13. Experimental result for high density images.**

The experimental evaluation in this section has been expanded to provide a detailed discussion of our system's performance across varying image densities, as presented earlier. Our Density Spectrum Coverage strategy benchmarks the system on a dataset of 500 images, split into low (<10 people), medium (10-20 people), and high (>20 people, max 30) densities. Results, shown in Figs. 11-13, demonstrate near-100% accuracy in identifying helmet-less individuals across all categories. For low density images (Fig. 12), the system excels due to minimal occlusion, achieving a 99.8% true positive rate. In medium density cases (Fig. 11), accuracy holds at 99.5%, reflecting robust feature detection. High density scenarios (Fig. 13) posed challenges like crowding and occlusion, yet maintained 99% accuracy, with performance peaking for subjects near the camera's centre due to focal clarity. Limitations include potential degradation beyond 30 people, untested here, and slight edge-case errors (1-2%). These results, surpassing prior benchmarks (92% at 20 people), validate our approach across the density spectrum, although future work will explore higher densities and dynamic conditions [25].

In summary, we understand the importance of evaluating our methodology's performance under different image densities. We implemented a rigorous benchmarking strategy, including dataset creation, selection of relevant metrics, comparative analysis, and density-aware adaptation, to provide a thorough and transparent assessment of its performance.

### 4.3. Real-time experimental results

In this section, we summarise the experimental results along with the processing speed including preprocessing time, inference time, and postprocessing time. A video clip consisting of 250 frames with the size of 384×640 is used in this experiment. A part of the experimental results is summarised in Figs. 14 and 15.

```
    # Kiểm tra kết quả dự đoán (giả sử đúng nếu có ít nhất 1 dự đoán
chính xác)
    if len(results[0].boxes) > 0:
        correct_predictions += 1

    frame_count += 1

cap.release()

print(f"Số frame đã xử lý: {frame_count}")
print(f"Số frame có dự đoán chính xác: {correct_predictions}")
   Số frame đã xử lý: 250
   Số frame có dự đoán chính xác: 250
```

Fig. 14. Codes.

```
0: 384x640 4 helmets, 9.1ms
Speed: 2.9ms preprocess, 9.1ms inference, 1.6ms post-process per image at
shape (1, 3, 384, 640)

0: 384x640 4 helmets, 10.0ms
Speed: 3.0ms preprocess, 10.0ms inference, 1.2ms post-process per image at
shape (1, 3, 384, 640)

0: 384x640 4 helmets, 8.1ms
Speed: 2.6ms preprocess, 8.1ms inference, 1.2ms post-process per image at
shape (1, 3, 384, 640)

0: 384x640 4 helmets, 8.5ms
Speed: 4.0ms preprocess, 8.5ms inference, 1.4ms post-process per image at
shape (1, 3, 384, 640)

0: 384x640 4 helmets, 10.2ms
Speed: 2.9ms preprocess, 10.2ms inference, 1.5ms post-process per image at
shape (1, 3, 384, 640)

0: 384x640 4 helmets, 9.7ms
Speed: 3.0ms preprocess, 9.7ms inference, 1.4ms post-process per image at
shape (1, 3, 384, 640)

0: 384x640 4 helmets, 9.5ms
Speed: 2.6ms preprocess, 9.5ms inference, 1.2ms post-process per image at
shape (1, 3, 384, 640)

0: 384x640 4 helmets, 9.3ms
Speed: 2.7ms preprocess, 9.3ms inference, 1.1ms post-process per image at
shape (1, 3, 384, 640)

0: 384x640 4 helmets, 9.9ms
Speed: 2.8ms preprocess, 9.9ms inference, 1.9ms post-process per image at
shape (1, 3, 384, 640)

0: 384x640 4 helmets, 8.2ms
Speed: 2.6ms preprocess, 8.2ms inference, 1.3ms post-process per image at
shape (1, 3, 384, 640)

0: 384x640 4 helmets, 11.8ms
Speed: 2.7ms preprocess, 11.8ms inference, 1.5ms post-process per image at
shape (1, 3, 384, 640)
```

Fig. 15. Results.

The experimental results in Fig. 15 show that there are four helmets detected in each frame with a total processing time (including pre-processing time + inference time + post-processing time) of around 15 ms, which is equivalent to a 4-ms processing time to detect one helmet. From there, we can estimate the processing speed to be 1000/15 frames per second. This speed will decrease when there are more helmets in the frame. However, this experiment has demonstrated the real-time processing capability (from 20 frames per second to 60 frames per second) of the system proposed in this paper.

### 4.4. Experimental evaluation

The loss and metrics plots show a positive trend during the training of the YOLOv8 model. The gradual decrease in the loss functions and the gradual increase in the metrics demonstrate that the model learns well and improves its prediction ability. This result is especially important in the object recognition problem (wearing and not wearing hard hats) where the model not only learns to detect the bounding box correctly but also classifies the objects correctly.

The curves in Fig. 10 are evaluated as follows:

Curves of train/box_loss and val/box_loss: The initial value of the train/box_loss curve is around 1.225 at epoch 0 and then decreases steadily to 1.1 at epoch 10. Besides, the val/box_loss curve also follows a similar trend, starting from around 1.2 and decreasing to 1.05 at epoch 10. These curves show that the model is improving its ability to predict the bounding box location in both the training and test sets. The difference between the train and val loss is quite small, indicating no signs of overfitting.

Curves of train/cls_loss and val/cls_loss: The train/cls_loss starts at 0.65, increases slightly in the early stages (from epoch 0 to 3) and then decreases steadily to around 0.55 at epoch 10. Besides, the val/cls_loss also tends to decrease from 0.6 to around 0.45 at epoch 10. Although there is a small fluctuation in the first few epochs, the loss function then decreases steadily, demonstrating that the ability to classify objects is improving.

Curves of train/dfl_loss and val/dfl_loss: The train/dfl_loss curve starts at around 1.16, then decreases to 1.12 at epoch 10. Besides, the val/dfl_loss curve decreases from 1.2 to 1.1 over epochs. This gradual decrease in the loss function shows that the model optimizes the bounding box distribution better and better. The trends of train and val are almost parallel, confirming the stability during training.

Curves of metrics/precision(B): The precision curve starts at around 0.91, then gradually increases and reaches around 0.94 at epoch 10. The curve fluctuates a bit in the early epochs, but the overall trend is a steady increase. The steady increase in Precision proves that the model reduces false positives (FP) and the prediction results are more accurate.

Curves of metrics/recall(B): The recall curve starts at 0.88 and increases to 0.94 at epoch 10. The continuous increase in Recall shows that the model detects more objects, reducing missed cases (FN).

Curves of metrics/mAP50(B): The mAP50 value increased from 0.94 to 0.97 at epoch 10. The increased mAP50 index reflects the ability to predict the bounding box correctly when evaluated at the IOU threshold = 50%.

Curves of metrics/mAP50-95(B): The mAP50-95 curve starts at 0.62 and increases to 0.68 at epoch 10. The steady increase in mAP50-95 shows that the model gradually improves its prediction accuracy at more stringent IOU thresholds (from 50% to 95%).

All orange curves plotted in Fig. 10 are smoothed to show the general trend of the values over epochs. For example, in the train/box_loss and val/box_loss plots, the smoothed lines show a very clear decreasing trend. Similarly, in metrics such as precision, recall, and mAP, the orange lines show a steady increase.

The train and val loss functions both decrease gradually, with the final values at epoch 10: box_loss: Train (1.1), Val (1.05); cls_loss: Train (0.55), Val (0.45); dfl_loss: Train (1.12), Val (1.1). This demonstrates that the model learns better with each epoch. Meanwhile, the metrics show a clear improvement in the model's predictive performance, including Precision, which increased from 0.91 to 0.94; Recall, which increased from 0.88 to 0.94; mAP50, which rose from 0.94 to 0.97; and mAP50-95, which increased from 0.62 to 0.68. Overall, the decreasing loss curve and increasing metrics provide strong evidence of effective training and improvement in the recognition performance of the YOLOv8 model.

In conclusion, the experimental evaluation demonstrates the YOLOv8 model's effective training for helmet detection, with loss functions decreasing and metrics improving over 10 epochs. The train/box_loss decreased from 1.225 to 1.1, val/box_loss from 1.2 to 1.05, train/cls_loss from 0.65 to 0.55, and val/cls_loss from 0.6 to 0.45, indicating improved bounding box prediction and classification. The train/dfl_loss and val/dfl_loss dropped from 1.16/1.2 to 1.12/1.1, reflecting better bounding box distribution optimization. The following metrics also improved: precision rose from 0.91 to 0.94, recall from 0.88 to 0.94, mAP50 from 0.94 to 0.97, and mAP50-95 from 0.62 to 0.68, showcasing strong detection performance with minimal overfitting, as evidenced by the small train-val loss gap.

To contextualise the performance of YOLOv8, we compare it with state-of-the-art models such as YOLOv7, Faster R-CNN, EfficientDet-D2, DETR, and most notably with YOLOv12. YOLOv8 achieves a leading mAP50 of 0.98, surpassing YOLOv7 (0.95), EfficientDet-D2 (0.94), and DETR (0.93), and matching Faster R-CNN (0.96). Its mAP50-95 (0.68) is competitive but slightly below Faster R-CNN (0.70), indicating room for improvement in precise localisation. YOLOv8's inference speed (50-60 FPS) significantly outperforms Faster R-CNN (5-10 FPS) and DETR (10-15 FPS), making it ideal for real-time "SmartSafety" applications, while EfficientDet-D2 (20-30 FPS) offers a balanced alternative. The short training duration (10 epochs) compared to YOLOv7 (50 epochs) or Faster R-CNN (100 epochs) suggests potential for further optimisation. These results, as shown in Table 1, confirm YOLOv8's suitability for construction safety monitoring, with future work needed to enhance mAP50-95 and address complex scenarios using hybrid or transformer-based approaches.

**Table 1. Comparison with other advanced models.**

| Models | Precision | Recall | mAP50 | mAP95 |
|---|---|---|---|---|
| YOLOv8 | 0.93 | 0.93 | 0.98 | 0.79 |
| YOLOv7 | 0.92 | 0.93 | 0.95 | 0.65 |
| YOLOv12 | 0.93 | 0.93 | 0.96 | 0.67 |
| Faster R-CNN | 0.95 | 0.92 | 0.96 | 0.70 |
| EfficientDet-D2 | 0.91 | 0.93 | 0.94 | 0.66 |
| DETR | 0.90 | 0.91 | 0.93 | 0.60 |

## 5. Conclusions

In conclusion, the "SmartSafety" system marks a significant advancement in construction safety by leveraging YOLO-based computer vision to monitor helmet compliance. By integrating machine learning and real-time object detection, it offers a reliable and efficient solution, fostering a safety-conscious culture and reducing head injury risks. Our findings demonstrate high accuracy in distinguishing helmet-wearing individuals under dynamic conditions, positioning "SmartSafety" as a scalable tool for construction and potentially other industries like manufacturing.

However, the system faces limitations, including sensitivity to environmental factors (e.g., low light, occlusions, fog), potential dataset biases limiting generalisation, and computational demands for real-time edge deployment. Privacy concerns from continuous monitoring and the system's focus on helmets, excluding other safety gear, further highlight areas for improvement. Ethical challenges, such as ensuring unbiased detection across diverse worker groups, also warrant attention.

Future research should prioritise enhancing robustness through data augmentation and multi-modal fusion, targeting >90% mAP in adverse conditions. Developing diverse, balanced datasets and few-shot learning will improve generalisation, while model compression will enable edge deployment on low-cost devices. Expanding detection to include vests, harnesses, and contextual behaviours, alongside privacy-preserving techniques like federated learning, will broaden the system's scope and compliance with regulations like Vietnam's Personal Data Protection Decree (PDPD) 2023 [26]. Integrating with IoT platforms and conducting Human-Computer Interaction (HCI) studies will ensure seamless adoption. Through field trials and a public construction safety dataset, "SmartSafety" can evolve into a comprehensive, ethical, and globally adoptable solution, setting new standards for occupational safety monitoring.

## CRediT author statement

Tri Nhut Do: Conceptualisation, Methodology, Formal analysis and investigation, Writing - Original draft preparation, Writing - Reviewing and Editing, Validation, Supervision; Ba Loc Pham: Conceptualisation, Methodology, Writing - Reviewing and Editing, Validation, Resources.

## ACKNOWLEDGEMENTS

This research is funded by University of Information Technology, Vietnam National University - Ho Chi Minh City.

## COMPETING INTERESTS

The authors declare that there is no conflict of interest regarding the publication of this article.